\pdfoutput=1

\documentclass[11pt]{article}

\usepackage[review]{EMNLP2022}

\usepackage{times}
\usepackage{latexsym}

\usepackage[T1]{fontenc}

\usepackage[utf8]{inputenc}

\usepackage{microtype}

\usepackage{inconsolata}
\usepackage{graphicx} 
\usepackage{algorithmic}
\usepackage{multirow}
\usepackage{amsmath}
\usepackage{tabularx}
\usepackage{subcaption}
\usepackage{booktabs}
\usepackage[linesnumbered,ruled,vlined]{algorithm2e}
\SetKwRepeat{Do}{do}{while}%

\title{TSMind: Alibaba and Soochow University's Submission to the WMT22 Translation Suggestion Task}
\author{Xin Ge$^1$\thanks{~~indicates equal contribution.}, Ke Wang$^{1*}$, Jiayi Wang$^1$, Nini Xiao$^{1,2}$, Xiangyu Duan$^2$, Yu Zhao$^1$, Yuqi Zhang$^1$\thanks{~~indicates the corresponding author.} \\
$^1$Alibaba Group Inc. \ \ \ $^2$Soochow University \\
\texttt{\{shiyi.gx,moyu.wk,joanne.wjy\}@alibaba-inc.com} \\ 
\texttt{\{nnxiaonnxiao,xiangyuduan\}@suda.edu.cn}, \texttt{\{kongyu, chenwei.zyq\}@alibaba-inc.com}
}

\begin{document}
\maketitle
\begin{abstract}
This paper describes the joint submission of Alibaba and Soochow University, TSMind, to the WMT 2022 Shared Task on Translation Suggestion (TS). We participate in the English $\leftrightarrow$ German and English $\leftrightarrow$ Chinese tasks. Basically, we utilize the model paradigm fine-tuning on the downstream tasks based on large-scale pre-trained models, which has recently achieved great success. We choose FAIR's WMT19 English $\leftrightarrow$ German news translation system and MBART50 for English $\leftrightarrow$ Chinese as our pre-trained models. Considering the task's condition of limited use of training data, we follow the data augmentation strategies proposed by \citet{yang2021wets} to boost our TS model performance. The difference is that we further involve the dual conditional cross-entropy model and GPT-2 language model to filter augmented data. The leader board finally shows that our submissions are ranked first in three of four language directions in the Naive TS task of the WMT22 Translation Suggestion task.
\end{abstract}

\section{Introduction}

Computer-aided translation (CAT) \citep{barrachina2009statistical,green2014predictive,green2015natural,knowles-koehn-2016-neural} has become more and more popular to help increase the quality of machine translation \citep{lopez2008statistical,koehn2009statistical} result. It also improves the efficiency of translators by combining the results of machine translation and the content edited by translators in the process of translation or post-editing \citep{bowker2002computer,lengyelmemoq,bowker2010computer,bowker2014computer,chatterjee2019automatic}. 

Post-editing based on machine translation is typical in CAT. Recent works \citep{domingo-etal-2016-interactive,gonzalez-rubio-etal-2016-beyond,peris2017interactive} propose interactive protocols and algorithms so that humans and machines can collaborate during translation, and machines can automatically provide feedback on humans' edits. One interesting mode is Translation Suggestion (TS) \citep{yang2021wets}, which offers alternatives for specific spans of words in the generated machine translation. It will be convenient if the model refines translation results in those specified locations with potential translation errors. \citet{yang2021wets} released a benchmark dataset for TS, \textit{WeTS}, which is one of the shared tasks in WMT22. At the same time, they proposed an end-to-end Transformer-like model for TS as the benchmark system. 

However, the lack of many labeled TS data limits the training of a large Transformer model to some extent. Though \citet{yang2021wets} have tried to utilize XLM-Roberta \cite{conneau2019unsupervised} to initialize the encoder of the Transformer, the decoder has to be trained from scratch, which leads to relatively low BLEU scores for some specific TS spans. We investigate the potential of other encoder-decoder pre-trained models by experiments to see if there is still room for improvement. Finally, we have found that pre-trained Transformer NMT models could be suitable choices to be fine-tuned with the limited size of TS data. In addition, we applied similar data augmentation strategies proposed in \citet{yang2021wets}, but use the well-trained alignment models between source and target languages from \citet{lu-etal-2020-alibaba} to filter out high-quality augmented data. Our submissions are ranked first in three of four language directions in the WMT22 Translation Suggestion task. 
\section{The Model}
\label{sec:model}
We train a simple end-to-end Transformer model for each language pair to generate the translation suggestion candidates. The source sentence and the masked translation, in which an incorrect span requiring an alternative has been replaced with a special mask tokens in advance, are concatenated with a special separation token \textit{[SEP]}. Afterward, we feed the concatenated sequence as input of the Transformer encoder and the translation suggestion needs to be generated by the Transformer decoder. The model is trained in the same way of a normal translation model.

Considering that the TS task also relies on alignments of hidden representations between the source and the target language, a well-trained translation model can be a good starting point for TS model training. The weights of our model are initialized with a pre-trained Transformer NMT model. Then, a two-phase training pipeline is applied. In the first phase, the model is trained with pseudo corpus derived from data augmentation described in Section \ref{sec:dataAug}. In the second phase, we fine-tune the model with the real TS train data released by the organizers.

\newcommand{\pluseq}{\mathrel{+}=}
\newcommand{\subeq}{\mathrel{-}=}

\begin{table}[t]
  \begin{tabularx}{1.0\linewidth}{c|X}
    \toprule
    Symbol & Definition\\
    \midrule
    $\mathbf{x}$ & Sentence in source language \\  \hline
    $\mathbf{y}$ & Machine translation result of $\mathbf{x}$ \\  \hline
    $\mathbf{r}$ & Reference sentence $\mathbf{x}$ \\ \hline
    $\mathbf{x}^i$ & The $i$-th token of $\mathbf{x}$ \\ \hline
    $\left\|\mathbf{x}\right\|$ & Length of $\mathbf{x}$, i.e. the number of tokens in $\mathbf{x}$ \\ \hline
    $\mathbf{x}^{i:j}$ & The fragment of ${\mathbf{x}}$ from position ${i}$ to ${j}$ \\  \hline
    $\mathbf{x}^{\neg i:j}$ & The masked version of $\mathbf{x}$, in which tokens at the position from ${i}$ to ${j}$ of $\mathbf{x}$ is replaced with a mask token. \\  \hline
    $\mathbf{{\hat{p}}}$ & All aligned-phrase pair between $\mathbf{y}$ and $\mathbf{r}$, pair look likes (${\mathbf{y}^{i:j}}$,  ${\mathbf{r}^{a:b}}$)  \\  \hline
    $\mathbf{\hat{y}}$ & Replace ${\mathbf{y}^{i:j}}$ with ${\mathbf{r}^{a:b}}$ in ${\mathbf{y}}$, and get another new sentence $\hat{\mathbf{y}}$ \\ 
    \bottomrule
\end{tabularx}
  \caption{Notations}
  \label{tab:notation}
\end{table}
\begin{table}[tbp]
\centering
\begin{tabular}{cccc}
\toprule
&WMT22& Filter Length& Filter Quality \\ 
\midrule
en-zh  & 23.2M         & 9.78M                 &    6.9M  \\ 
\midrule
en-de  & 30.0M    & 12.73M        & 8.18M                  \\
\bottomrule
\end{tabular} 
\caption{Number of parallel samples remained after filtering by length and cross-entropy quality score \cite{lu-etal-2020-alibaba}.}
\label{tab:parallel_corpus_filter}
\end{table}

\begin{figure*}[h]
\centering
\includegraphics[width=0.9\linewidth]{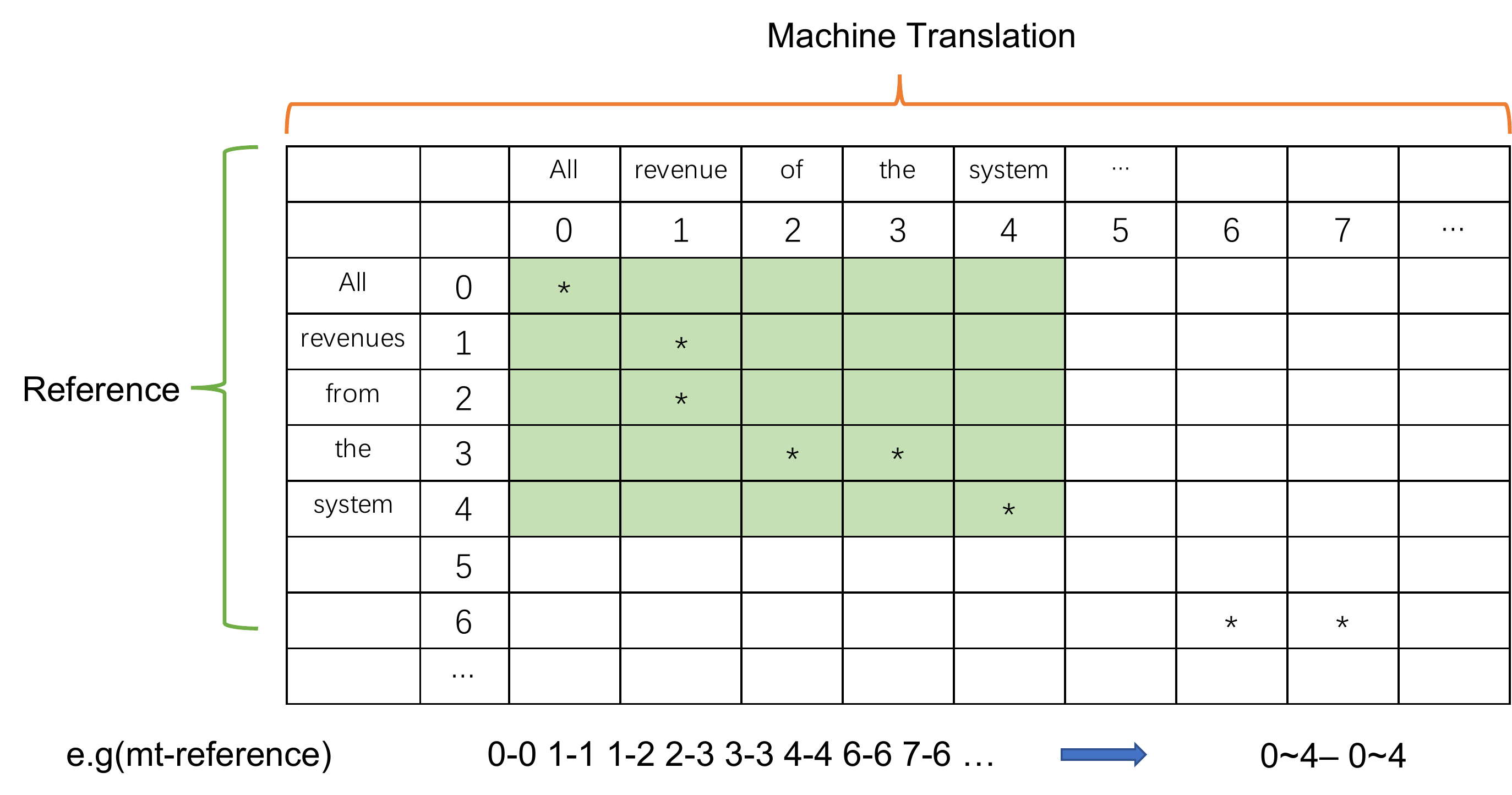} 
\caption{In this example, we have the alignment info between machine translation (MT) and reference sentences: $0\text{-}0$, $1\text{-}1$, $1\text{-}2$, $2\text{-}3$, $3\text{-}3$ $4\text{-}4$, $6\text{-}6$, $6\text{-}7$, the phrase from $0\sim4$ in MT are aligned to $0\sim4$ in reference. The rectangle enclosed by the aligned phrases between MT and reference should satisfy that each row and each column has at least one *.}
\label{fig:fast-align}
\end{figure*}

\begin{figure*}[htp]
\centering
\includegraphics[width=0.9\linewidth]{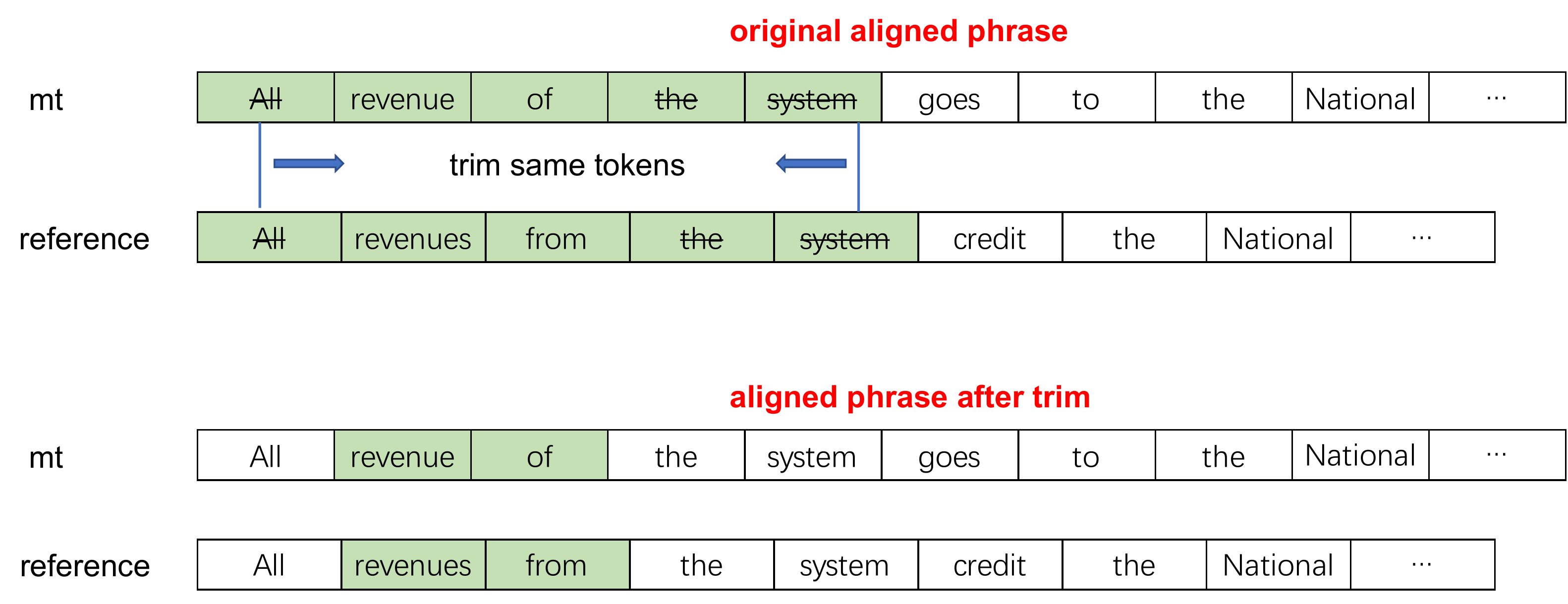} 
\caption{As shown in Figure \ref{fig:fast-align}, we get the original aligned phrase between MT and reference which are "All revenue of the system" and "All revenues from the system". We then trim the tokens that appear in both MT and reference to compress the aligned phrase. Finally, we get the trimmed aligned phrase: "revenue of" and "revenues from"}
\label{fig:trim-align}
\end{figure*}

\section{Data Augmentation}
\label{sec:dataAug}
 We follow the data augmentation methods provided by \citep{yang2021wets} to generate three types of pseudo data for TS model training: 
data sampled on the golden parallel corpus, data sampled on the pseudo parallel corpus, and data extracted with word alignment. 
However, the details of the pseudo data augmentation in this paper are slightly different from those of \citet{yang2021wets}. Full details are exhibited in the following subsections.

\begin{algorithm}
\caption{Algorithm of Phrase Align}
\label{alg:phrase-align}
\SetKwInput{KwInput}{Input}                
\SetKwInput{KwOutput}{Output}              
\DontPrintSemicolon
  
  \KwInput{$\mathbf{y}$, $\mathbf{r}$, $\mathbf{A}$}
  \KwOutput{$\hat{\mathbf{p}}$}

  \SetKwFunction{IsMatch}{IsMatch}
  \SetKwFunction{GenerateAlign}{GenerateAlign}
 
  \SetKwProg{Fn}{Function}{:}{}
  \Fn{\GenerateAlign{$\mathbf{y}$, $\mathbf{r}$, $\mathbf{A}$}}{
     $yt = size(\mathbf{y})$, $rt = size(\mathbf{r})$ \\
    \For{$i\gets0$ \KwTo $yt$}{
        \For{$j\gets i$ \KwTo $yt$}{
            \For{$a\gets0$ \KwTo $rt$}{
                \For{$b\gets a$ \KwTo $rt$}{
                    \uIf{\textit{IsMatch}($\mathbf{y}$, $\mathbf{r}$,$i$, $j$, $a$, $b$, $\mathbf{A}$)}{
                        \Do{ $\mathbf{y}^{i} == \mathbf{r}^{a}$}{
                            $i\pluseq1$;
                            $a\pluseq1$
                        }
                        \Do{ $\mathbf{y}^{j} == \mathbf{r}^{b}$}{
                            $j\subeq 1$;
                            $b\subeq 1$
                        }
                        $\hat{\mathbf{p}}.add((\mathbf{y}^{i:j}, \mathbf{r}^{a:b}))$
                    }
                }
            }
        }
    }
    \KwRet $\hat{\mathbf{p}}$
  }
  
  \SetKwProg{Fn}{Function}{:}{\KwRet}
  \Fn{\IsMatch{$\mathbf{y}$, $\mathbf{r}$, $i$,$j$,$a$,$b$,$\mathbf{A}$}}{
        \For{$ii\gets i$ \KwTo $j$}{
            \textbf{let} T = \{$t_i$| $\mathbf{r}^{t_i}$\ is\ aligned\ with\ $\mathbf{y}^{ii}$\ in\ $\mathbf{A}$ \}
            \ForEach{$t_i \in T$}{%
              
              \If{$t_i \textless a$ or $t_i \textgreater b$}{%
                \KwRet \textbf{False}
              }
            }
        }
        
        \For{$aa\gets a$ \KwTo $b$}{
        \textbf{let} T $=$ \{$t_a$| $\mathbf{r}^{aa}$\ is\ aligned\ with\ $\mathbf{y}^{t_a}$\ in\ $\mathbf{A}$ \}
            \ForEach{$t_a \in T$}{%
              
              \If{$t_a \textless i$ or $t_a \textgreater j$}{%
                \KwRet \textbf{False}
              }
            }
        }
        \KwRet \textbf{True}
  }
\end{algorithm}

\subsection{Sampling from golden parallel corpus}
\label{golden_corpus}

Raw parallel corpus is firstly filtered by the sentence length. All sentence pairs that have less than 20 words or more than 80 words on any side are removed. 

Considering that there might be noise data in the corpus, we apply the dual conditional cross-entropy model \cite{lu-etal-2020-alibaba} to obtain a quality score for each sample. Sentence pairs with low quality are filtered.

Then we generate a pseudo corpus with the remained high-quality parallel corpus. ($\mathbf{x}$, $\mathbf{r}$) is marked as the sentence pair of the parallel corpus, where $\mathbf{x}$ is the source sentence and $\mathbf{r}$ is the golden reference. $\left\|\mathbf{r}\right\|$ represents the number of tokens in $\mathbf{r}$. 

The first step is to randomly sample the length ${l}$ to mask for the reference ${r}$ from a uniform distribution: 
\begin{equation}
\label{eq:mask_length}
    l \sim \textit{U}(1,\left\|\mathbf{r}\right\|) 
\end{equation}
Then a span with $l$ tokens $\mathbf{r}^{i:j}$ is randomly selected by: 
\begin{equation}
\label{eq:mask_position}
    i \sim \textit{U}(0,\left\|\mathbf{r}\right\| - l),\ \ 
    j = i+l
\end{equation}


Finally, we get the TS training data $(\mathbf{x}, {\mathbf{r}^{\neg i:j}}, {r^{i:j}})$ from each parallel sentence pair $(\mathbf{x}, \mathbf{r})$, where ${\mathbf{r}^{\neg i:j}}$ is denoted as the masked version of $r$, in which ${\mathbf{r}^{i:j}}$ is replaced with a mask token, e.g <MASK\_REP>.

\subsection{Sampling on Pseudo Parallel Corpus}
\label{pseudo_corpus}
In addition, the monolingual corpus is another source for data augmentation. We first filter the monolingual data with a language identification process. Then pseudo parallel corpus is generated with NMT models. Finally, TS training data can be generated as we do in Section \ref{golden_corpus}. 



\begin{table*}[h]
\centering
\begin{tabular}{lcccc}
\hline
            Method               & En-De & De-En & En-Zh & Zh-En \\ \hline
TSMind                     & 45.90 & 43.37 & 30.21 & 28.77 \\
-w/o first-phase training  & 37.14 & 33.23 & 21.20 & 16.44 \\
-w/o second-phase training & 37.37 & 36.83 & 21.84 & 19.19 \\ \hline
\end{tabular}
\caption{Sacre-BLEU on the validation sets of Sub-Task 1 (Naive TS) of the WMT'22 Translation Suggestion Task.}
\label{tab:devRes}
\end{table*}

\subsection{Extracting with Word Alignment}
\label{align_corpus}

In the task of TS, the labels for the masked span is always correct while the translation contexts of the span, $\mathbf{y}^{\neg i:j}$ are not error-free. Therefore, both of the above two types of pseudo data are biased from the task. In pseudo data sampled from golden parallel corpus, the translation contexts are error-free. And the labels of pseudo data from machine translation results are not always correct. To reduce the bias, another way of data augmentation is proposed in \citet{yang2021wets}. They utilize the alignment between the machine translation and the golden reference to generate pseudo-training samples for TS. We use the similar idea and the details of our alignment-based data augmentation algorithm are described as follows.

Given the triplet ($\mathbf{x}$, $\mathbf{y}$, $\mathbf{r}$) where $\mathbf{x}$ is the source sentence, $\mathbf{y}$ is the machine translation result generated by NMT models, and $\mathbf{r}$ is the reference, we need to find aligned segment pairs $(\mathbf{y}^{i:j}, \mathbf{r}^{a:b})$ between $\mathbf{y}$ and $\mathbf{r}$. 

First, we use the Fast Align toolkit \cite{dyer2013simple} to extract token alignments between $\mathbf{y}$ and $\mathbf{r}$.
The align result $\mathbf{A}$ is a list of aligned indexes in the format of $i\text{-}a$, which means token $\mathbf{y}^i$ is aligned to $\mathbf{r}^a$. 
With the token alignments, the next step is to extract aligned-phrase pairs, denoted as ${\mathbf{\hat{p}}}$. 
Figure \ref{fig:fast-align} shows an example of an aligned phrase between MT and reference. 
The algorithm of the aligned-phrase extraction is presented in Algorithm  \ref{alg:phrase-align} from line $1$ to line $13$. 
The aligned phrases are a subset of SMT's phrase extraction \cite{koehn-etal-2003-statistical} with two restricts. 1) Each row and each column of a aligned phrase has at least one token aligned (a * in Figure \ref{fig:fast-align}); 2) We take only the longest phrase and the sub-phrases are not taken. 
After the original aligned phrase is obtained, we remove tokens that appear in both MT and reference to get the trimmed result as shown in Figure \ref{fig:trim-align}. We trim these common tokens because we want the model to focus more on the incorrect spans and its alternatives. The pseudo-code of the phrase-alignment is presented in the Algorithm \ref{alg:phrase-align}. We denote the aligned phrase as $\mathbf{y}^{i:j}$ and $\mathbf{r}^{a:b}$, $\mathbf{y}^{\neg i:j}$ represents the masked version of $\mathbf{y}$ as described in Section \ref{pseudo_corpus}. 

Now we need to judge whether $\mathbf{r}^{a:b}$ is better than $\mathbf{y}^{i:j}$ in the context of $\mathbf{y}^{\neg i:j}$. We replace $\mathbf{y}^{i:j}$ with $\mathbf{r}^{a:b}$ in $\mathbf{y}$, and get another new sentence $\mathbf{\hat{y}}$. First, we use the dual conditional cross-entropy model as described in Section \ref{golden_corpus} to calculate the quality score of $(\mathbf{x}, \mathbf{\hat{y}})$. Then, the perplexity of $\mathbf{\hat{y}}$ and $\mathbf{y}$ are given by the language-specific GPT2 models \cite{stefan_schweter_2020_4275046, radford2019language,zhao2019uer} released on HuggingFace \cite{wolf-etal-2020-transformers} respectively. 
If the cross-entropy quality score of $(\mathbf{x}, \mathbf{\hat{y}})$ is smaller than the threshold of $\beta_1$ 
and the perplexity loss reduction value of $\mathbf{y} - \mathbf{\hat{y}}$ is at least $\beta_2$, then the translation $\mathbf{\hat{y}}$ is most likely better than $\mathbf{y}$. We can treat $\mathbf{y}^{\neg i:j}$ as the masked version of MT and $\mathbf{r}^{a:b}$ as the correct alternative. $\beta_1$ and $\beta_2$ are the hyper-parameters of the alignment.

Finally, we get the aligned training data ($\mathbf{x}$, ${\mathbf{y}^{\neg i:j}}$, ${\mathbf{r}^{a:b}}$) from the triplets $(\mathbf{x}, \mathbf{y}, \mathbf{r})$.

\begin{table*}[]
\centering
\begin{tabular}{rccccc}
\hline
                        & En-De & De-En & En-Zh & Zh-En & Average \\ \hline
XLM-R                   & 25.12 & 27.40 & 32.48 & 21.25 & 26.56   \\
Naïve Transformer       & 28.15 & 30.08 & 35.01 & 24.20 & 29.36   \\
Dual-source Transformer & 28.09 & 30.23 & 35.10 & 24.29 & 29.43   \\
SA-Transformer          & 29.48 & 31.20 & \textbf{36.28} & 25.51 & 30.62   \\
TSMind                  & \textbf{47.44} & \textbf{45.02} & 26.41 & \textbf{31.78} & \textbf{37.66}   \\ \hline
\end{tabular}
\caption{Sacre-BLEU on the test sets of WeTS \cite{yang2021wets}}
\label{tab:weTStestRes}
\end{table*}


\section{Experiment}
\subsection{Corpus and Setup}
Parallel corpora for data augmentation in Section \ref{golden_corpus} and \ref{align_corpus} and monolingual corpora for Section \ref{pseudo_corpus} are all downloaded from WMT22 general translation task\footnote{https://statmt.org/wmt22/translation-task.html}. For English $\leftrightarrow$ German, WikiMatrix \cite{schwenk-etal-2021-wikimatrix}, News Commentary v16, Common Crawl Corpora, and Tilde MODEL Corpora \cite{rozis-skadins-2017-tilde} are used as parallel corpus. For English $\leftrightarrow$ Chinese, parallel corpus we used includes UN Parallel Corpus V1.0 \cite{ziemski-etal-2016-united} and all parallel corpora from CCMT corpus \cite{yang2019ccmt} except for the casict2015 corpora.
For monolingual corpora, News Commentary and News Crawl are used for all three languages, and Leipzig Corpora \cite{goldhahn-etal-2012-building} is also used for Chinese and German.

Then the filtering strategies proposed in Section \ref{golden_corpus}
are applied to the raw parallel data. 
The number of data remained after every filtering step can be found in Table \ref{tab:parallel_corpus_filter}.

We download monolingual data from WMT22, and get a total of 45.02 million German, 14.68 million English and 10.01 million Chinese monolingual sentences.

For data augmentation in Sections \ref{pseudo_corpus} and \ref{align_corpus}, we use the NMT models for English $\leftrightarrow$ German and English $\leftrightarrow$ Chinese released by \citet{yang2021wets}\footnote{https://github.com/ZhenYangIACAS/WeTS} to translate the source sentences. 
And the hyper-parameter $\beta_1$ and $\beta_2$ to filter aligned phrases are set to 2.5 and 0.05, respectively. 

\subsection{Model Training}

As mentioned in Section \ref{sec:model}, a well-trained NMT model is a good starting point for the TS model. For English $\leftrightarrow$ German, we initialize the weights with the NMT models released by \citet{ng2019facebook} (Winner of WMT'19). For English $\leftrightarrow$ Chinese, the one-to-many and many-to-one mBART50 models \citep{tang2020multilingual} are used.

We use the fairseq toolkit \cite{ott2019fairseq} to train and evaluate our model. Hyper-parameters are set to the same as examples in the fairseq toolkit 
except that we reset the learning rate at the beginning of the first phase training and beam size is set as 6 during inference.


\subsection{Experimental Results}
We evaluate the TSMind by calculating the Sacre-BLEU \citep{post2018call} of the top-$1$ generated translation suggestion candidate on the golden reference. Results of the validation sets of WMT22 are shown in Table \ref{tab:devRes}. Without first-phase training, we get much worse performances. This demonstrates that a large amount of pseudo corpora contributes much to the model. However, without the second-phase training (i.e. without the human-labeled data), we cannot obtain a good translation suggestion model with only pseudo corpora either. Therefore, the design of the two-phase training and the pseudo corpora are essential to set good translation suggestions. 

Since the development set of WMT'22 is not the same as the test set used in  \citet{yang2021wets}, to make a fair comparison, we also report the Sacre-BLEU on the test set of WeTS in Table \ref{tab:weTStestRes}. Results of all baseline systems are reported by \citet{yang2021wets}. TSMind outperforms the strong baseline, SA-Transformer, significantly with a gap of 7.04 BLEU on average for all four language pairs. We notice that TSMind does not perform well on the English to Chinese language pair. The reason might be that the pre-trained model we use is the one-to-many model of mBART50, and the multilingual decoder is not well-trained for Chinese. For example, on the English to Chinese news translation test set of WMT'20 \cite{barrault-etal-2020-findings}, mBART50 only achieves a Sacre-BLEU value of 30.79, while the Sacre-BLEU of state-of-the-art is 49.2.


\section{Conclusion}
In this paper, we present our translation suggestion systems, TSMind, for the WMT 2022 Translation Suggestion Task. Different from previous work, we use well-trained NMT models as the pre-trained models and applied a two-phase training strategy. 

We explore three data augmentation strategies from previous work and utilize the dual conditional cross-entropy model to filter out low-quality augmented data. The leader board finally shows that our submissions are ranked first in three of four language directions in the Naive TS task of WMT22 Translation Suggestion task.
\bibliography{anthology,custom}
\bibliographystyle{acl_natbib}




\end{document}